%% file: variational_variance_main.tex
\documentclass[twoside]{article}

\usepackage[preprint]{aistats2021}
\usepackage[round]{natbib}

\usepackage{amsmath,amssymb,bm}
\usepackage{booktabs}
\usepackage{hyperref,cleveref}
\usepackage{float,graphicx}
\usepackage{tikz}
\usetikzlibrary{bayesnet}
\usepackage{subfig}

\DeclareMathOperator{\D}{\mathcal{D}}

\DeclareMathOperator*{\E}{\mathbb{E}}
\DeclareMathOperator{\ELBO}{\mathcal{L}}
\DeclareMathOperator{\Gam}{\text{Gamma}}
\newcommand{\iid}{\overset{iid}{\sim}}

\DeclareMathOperator{\N}{\mathcal{N}}

\DeclareMathOperator{\T}{\text{T}}
\DeclareMathOperator{\Uni}{\text{Uniform}}
\DeclareMathOperator{\var}{\text{var}}

\newcommand{\DetCode}{\href{https://github.com/SkafteNicki/john}{code}}

\newcommand{\meanpmstd}{(mean$\pm$std.)}
\newcommand{\tuple}{Tuples below dataset are $(N_{\text{observations}},\dim(x),\dim(y))$.}

\begin{document}

% If your paper is accepted and the title of your paper is very long,
% the style will print as headings an error message. Use the following
% command to supply a shorter title of your paper so that it can be
% used as headings.
%
%\runningtitle{I use this title instead because the last one was very long}

% If your paper is accepted and the number of authors is large, the
% style will print as headings an error message. Use the following
% command to supply a shorter version of the authors names so that
% they can be used as headings (for example, use only the surnames)
%
%\runningauthor{Surname 1, Surname 2, Surname 3, ...., Surname n}

\twocolumn[

\aistatstitle{Variational Variance: Simple, Reliable, Calibrated \\ Heteroscedastic Noise Variance Parameterization}

\aistatsauthor{ Andrew Stirn \And David A. Knowles }

\aistatsaddress{
	Columbia University \\ andrew.stirn@cs.columbia.edu
	\And
	Columbia University \& New York Genome Center \\
	daknowles@cs.columbia.edu} ]

\begin{abstract}
Brittle optimization has been observed to adversely impact model likelihoods for regression and VAEs when simultaneously fitting neural network mappings from a (random) variable onto the mean and variance of a dependent Gaussian variable. Previous works have bolstered optimization and improved likelihoods, but fail other basic posterior predictive checks (PPCs). Under the PPC framework, we propose critiques to test predictive mean and variance calibration and the predictive distribution's ability to generate sensible data. We find that our attractively simple solution, to treat heteroscedastic variance variationally, sufficiently regularizes variance to pass these PPCs. We consider a diverse gamut of existing and novel priors and find our methods preserve or outperform existing model likelihoods while significantly improving parameter calibration and sample quality for regression and VAEs.
\end{abstract}

\section{Introduction}
\label{sec:intro}
The machine learning community ubiquitously employs neural networks to map conditioning (random) variables onto the parameter space of other model variables. This technique leverages the expressive power of deep learning while preserving probabilistic interpretability. For example, we often map covariates onto the simplex with neural networks to parameterize a categorical distribution over observed labels in classification. Parameterizing the mean and variance of a normal distribution with neural networks is also prevalent \citep{nix1994estimating,kingma2013auto,rezende2014stochastic} but problematic. In particular, if our conditional mean network predicts nearly perfectly (i.e. $\mu(x_i)\approx y_i \ \forall \ i\in[N]$), then maximizing the log likelihood will push the variance network $\sigma^2(x_i)$ towards a pathological 0. This tendency coupled with the fact that $\sigma^{-2}$ (precision) appears as a multiplicative factor in the gradient of the normal log likelihood w.r.t. $\mu$, underlies why jointly optimizing mean and variance networks can be unstable. As the mean estimates $\mu(x_i)$ improve, the log likelihood encourages $\sigma^{-2}(x_i)\rightarrow\infty$ such that minuscule errors by the mean network can produce inappropriately large parameter updates. The variance network $\sigma^2(x_i)$ effectively controls the learning rate of the mean network $\mu(x_i)$--increasing it as the mean network improves--in direct opposition to the stochastic gradient descent convergence criteria of \cite{robbins1951stochastic}. While good optima can be found when these criteria are not met, instability has been observed to reduce model likelihoods when optimizing mean and variance networks in regression \citep{detlefsen2019reliable} as well as when using mean and variance networks for Gaussian decoder likelihoods in VAEs \citep{takahashi2018student}. Also, \cite{dai2018diagnosing} identify the theoretical preference of an optimal decoder for zero variance.

While substantial progress has been made in producing accurate means of posited model densities, reliable variance estimation has been more elusive despite its critical importance for numerous applications in today's machine learning arena. For regression, accurate variance estimates enable Bayesian \emph{active learning} \citep{cohn1996active} and \emph{reinforcement learning} \citep{ghavamzadeh2016bayesian} regimes where new data is requested or exploration carried out based on predictive variance. Realistic sample generation requires well-calibrated variance and is critical to \emph{adversarial learning} and \emph{data imputation}. We carefully inspected the code of many state-of-the-art VAE methods and found that the generated `samples' were rarely sampled from the predictive (decoder) distribution. Instead, these methods ancestrally resample latent variables from the variational posterior and report the expected value of the decoder density. This `sampling' procedure is actually a Monte-Carlo estimate--often using just a single sample from $q(z|x)$--of the posterior predictive mean, $\E[x^* | x] \triangleq \int x^* p(x^*|z)  q(z|x) \ dz$. While preserving uncertainty on the latent space by sampling $q(z|x)$, reporting this expectation over the observed data space obfuscates any uncertainty in the predictive density. Furthermore, approximating the posterior predictive distribution, $p(x^*|x) \triangleq \int p(x^*|z) q(z|x) \ dz$, with too few $z$ samples can lead to inaccurate predictive variance estimates. VAE papers claiming improvements to sample quality \citep{van2017neural,razavi2019generating} and imputation \citep{nazabal2018handling,mattei2018miwae} do \emph{not} sample the predictive distribution despite sometimes fitting a global (homoscedastic) scalar variance to improve mean calibration \citep{dai2018diagnosing}.

Perhaps the two most commonly reported performance metrics for regression and VAEs are the log likelihood and the root mean square error (RMSE) of the model's expected value. In VAE papers, it is not always clear if the reported log likelihood is the \emph{expected log likelihood}, $\E_{q(z|x)}[\log p(x^*|z)]$, from the variational objective or the \emph{log posterior predictive probability}, $\log p(x^*|x) = \log \E_{q(z|x)}[p(x^*|z)]$, where $x^*$ is a replicated $x$. The former is a lower bound of the latter via Jensen's inequality. For consistency, we always use the (posterior) predictive distributions, $p(y|x)$ for regression and $p(x^*|x)$ for VAEs, since they are equally well-defined for the frequentist and Bayesian methods we consider. We too evaluate log likelihoods and mean RMSEs, but also look beyond to variance. To do so, we adopt the framework of posterior predictive checks (PPCs) \citep{gelman2013bayesian}. PPCs posit a well-fit model should, with high probability, produce new data that looks similar to the observed data since any discrepancy could be the result of model misfit or chance. A common PPC is to evaluate the posterior predictive likelihood on a replicated set of the training data or better yet on a held-out validation or test set. Alternatively, one can sample values from the predictive distribution and look for systemic discrepancies with the original data that may indicate model failure. Here, we conduct PPCs from several perspectives. We assess mean and variance calibrations by measuring bias and RMSE between the predictive mean and the data and between the predictive variance and the empirical variance w.r.t. the predictive mean. We also measure bias and RMSE between the original data and samples taken from the predictive distribution; this tests the model's ability to generate sensible data thereby critiquing cooperation of the mean and variance. We provide PPC specifics for regression and VAEs respectively in \cref{sec:regression,sec:vae}.

In this article, we advocate a Bayesian treatment of the predictive distribution's variance (or rather precision for computational convenience). Treating variance variationally induces a Kullback–Leibler (KL) divergence, which, for an appropriate prior, will produce gradients that prohibit variance from approaching the aforementioned zero pathology (somewhat analogous to logarithmic barriers to enforce constraints in convex optimization) and alleviates the theoretical concern that maximum likelihood estimation (MLE) of continuous VAEs is ill-posed for unbounded likelihood functions \citep{mattei2018leveraging}. \cite{detlefsen2019reliable} and \cite{takahashi2018student} have addressed optimization instabilities to improve regression and VAE model likelihoods, respectively. Here, we look beyond likelihood to holistically improve model performance through the lens of PPCs. In \cref{sec:vi}, we review relevant variational inference concepts. In \cref{sec:regression}, we formalize our proposed methods for regression and experimentally compare our methods to a variety of baselines. In \cref{sec:vae}, we do the same but for VAEs. We emphasize that our proposals broadly apply to both regression and continuous VAEs and notably outperform other methods specific to each context.

\section{Amortized Variational Inference}
\label{sec:vi}
Variational inference (VI) \citep{blei2017variational} posits a family of tractable distributions $q(\Theta;\nu)$ to approximate the true posterior $p(\Theta | \D)$ over latent variables $\Theta$ conditioned on observed data $\D$. We assume i.i.d. data such that $p(\D|\Theta)=\prod_{d\in\D}p(d|\Theta)$. Often, and as in the case of \emph{amortized inference} \citep{kingma2013auto}, we use a neural network with shared learnable parameters $\phi$ to map data $d$ onto the variational parameters $\nu$ (i.e. $f_\phi: d \rightarrow \nu$). Amortized VI minimizes the variational posterior's KL divergence from the true posterior, $D_{KL}(q(\Theta|\D) \ || \ p(\Theta | \D))$, by maximizing the evidence lower bound (ELBO or $\ELBO$ or short),
\begin{align*}
    \sum_{d\in\D}\E_{q(\Theta|f_\phi(d))}\big[\log p(d|\Theta)\big] - D_{KL}\big(q(\Theta|f_\phi(d)) \ || \ p(\Theta)\big),
\end{align*}
since KL divergence is strictly non-negative and the summation of these dual objectives equals a constant. We focus on Gaussian likelihoods with mean $\mu$ and precision $\lambda$. If we treat $\lambda$ variationally--we consider it a latent variable (i.e. $\lambda \in \Theta$), specify a prior $p(\lambda)$ to describe its generative process, and employ a variational family $q(\lambda|\cdot)$ to approximate the posterior (or factor thereof)--then the KL divergence above will contain $D_{KL}(q(\lambda|\cdot) \ || \ p(\lambda))$. This regularizing term can fortify optimization as discussed in \cref{sec:intro} and, with well-informed priors, ideally will find distributions over $\lambda$ that accurately reflect the local predictive ability of mean network $\mu(\cdot)$.

\begin{figure}[t]
\centering
\begin{tikzpicture}
    \node[obs] (x) {$x_i$};
    \node[obs, below=2cm of x] (y) {$y_i$};
    \node[det, below=0.5cm of x, xshift=-0.45cm] (m) {$\mu_i$};
    \node[det, below=0.41cm of x, xshift=0.45cm] (s) {$\sigma^2_i$};
    \edge {x} {m,s};
    \factor[above=of y] {w-f} {left:$\N$} {m,s} {y};
	\plate {p} {(x)(m)(s)(y)} {$i \in [N]$};
\end{tikzpicture}
\begin{tikzpicture}
    \node[obs] (x) {$x_i$};
    \node[obs, below=2cm of x] (y) {$y_i$};
    \node[det, below=0.5cm of x, xshift=-0.8cm] (m) {$\mu_i$};
    \node[det, below=0.5cm of x, xshift=0cm] (a) {$\alpha_i$};
    \node[det, below=0.46cm of x, xshift=0.8cm] (b) {$\beta_i$};
    \edge {x} {m,a,b};
    \factor[above=of y] {n} {left:$t$} {m,a,b} {y};
	\plate {p} {(x)(m)(a)(b)(y)} {$i \in [N]$};
\end{tikzpicture}
\begin{tikzpicture}
    \node[obs] (x) {$x_i$};
    \node[obs, below=2cm of x, xshift=-0.4cm] (y) {$y_i$};
    \node[det, below=0.5cm of x, xshift=-0.8cm] (m) {$\mu_i$};
    \node[det, below=0.5cm of x, xshift=0cm] (a) {$\alpha_i$};
    \node[det, below=0.46cm of x, xshift=0.8cm] (b) {$\beta_i$};
    \node[latent, below=2cm of x, xshift=0.4cm] (l) {$\lambda_i$};
    \factor[right=0.7cm of l] {p} {right:$\Gamma$} {} {l};
    \factor[above=of l] {g} {right:$\Gamma$} {} {};
    \draw[-] (a) -- (g) [dashed];
    \draw[-] (b) -- (g) [dashed];
    \draw[->] (g) -- (l) [dashed];
    \edge {x} {m};
    \draw[->] (x) -- (a) [dashed];
    \draw[->] (x) -- (b) [dashed];
    \factor[above=of y] {n} {left:$\N$} {m,l} {y};
	\plate {p} {(x)(m)(a)(b)(y)} {$i \in [N]$};
\end{tikzpicture}
\caption{Graphical Models for Regression: Normal, Student's $t$, and Variational Variance (left to right). Diamonds are deterministic neural network parameter maps. Solid arrows denote the generative process. Dashed arrows define the variational family.}
\label{fig:regression}
\end{figure}
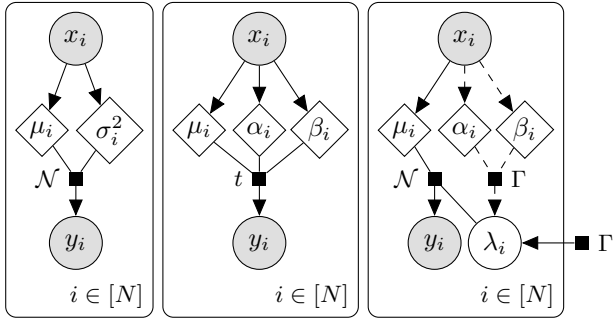

\section{Variational Variance for Regression}
\label{sec:regression}
Homoscedastic regression  assumes $y_i = f(x_i) + \epsilon_i$ where $f(x_i)$ is some unknown function of the covariates and $\epsilon_i \iid \N(0,\sigma^2)$ captures noise on $y_i$ via the unknown global variance parameter $\sigma^2$. Notably, the noise $\epsilon_i$ has no dependence on the covariates $x_i$. Maximizing the log likelihood of the data under a normal treatment can be accomplished by first minimizing $\sum_{i=1}^N (y_i - \mu(x_i))^2$ w.r.t. $\mu(\cdot)$ (in our case, a neural network) and thereafter setting the global noise variance $\sigma^2 = N^{-1} \sum_{i=1}^N (y_i - \mu(x_i))^2$. Unfortunately, homoscedasticity is often assumed out of convenience rather than to reflect prior knowledge of the true generative process. One can introduce heteroscedasticity by additionally parameterizing the noise variance with a neural network $\sigma^2(\cdot)$ operating on the covariates. The generative process for this model appears in the leftmost model of \cref{fig:regression}, has local likelihood $p(y_i|x_i)\triangleq \N(y_i|\mu(x_i),\sigma^2(x_i))$, can be optimized by MLE, and yields a predictive distribution that is simply the factorized normal likelihoods since there are no priors. We refer to this model as the \textbf{Normal} model and note $\sigma^2(\cdot)$ applies a softplus as its final activation to ensure positive variances.

\subsection{Existing Methods We Use as Baselines}
\label{sec:detlefsen}
In addition to optimization instabilities, \cite{detlefsen2019reliable} observed that simultaneously learning neural mean and variance parameterizations can also underestimate the true variance, especially in areas of covariate space with scarce data. They motivate and propose four tricks to ameliorate these issues. First, they argue a batch containing $x_i$, but lacking other nearby data, while sufficient for updating the mean, is insufficient for updating the variance (unless the mean is already known). Accordingly, they propose a `locality sampler' that ensures any batch sample $(x_i,y_i)$ is accompanied by its $K$ nearest neighbors (w.r.t. $x_i$), which are found in pre-training. Unfortunately, nearest neighbor distance can produce meaningless relationships for high dimensional data such as natural images. Second, they optimize the mean and variance networks in isolation (analogous to coordinate ascent). The first half of training fits only the mean network (using a fixed variance) to ensure that, during the latter half of training where coordinate ascent alternates every few batches, variance estimation is feasible since the mean network is presumably now reasonable. Third, they utilize a Gamma-Normal parameterized Student's $t$, $\T(y_i|x_i)\equiv \int_0^\infty \N(y_i|\mu(x_i),\lambda_i)\Gam(\lambda_i|\alpha(x_i),\beta(x_i))d\lambda_i$, as the likelihood, which again results in a predictive distribution that factors into local likelihoods (\cref{fig:regression}, middle). This parameterization highlights that the Student's $t$ distribution is a scaled mixture of Gaussians with unknown precision $\lambda_i$. From one perspective, the Student's $t$ regression model is an MLE problem unto itself. Alternatively, one could consider it an Empirical Bayes MAP (maximum a posteriori) estimation with local likelihood $\N(y_i|\mu(x_i),\lambda_i)$ and local heteroscedastic precision prior $\Gam(\lambda_i|\alpha(x_i),\beta(x_i))$, where the shared $\alpha(\cdot)$ and $\beta(\cdot)$ neural prior parameterizations are fit during inference. Lastly, they extrapolate variance as a learnable convex combination between the estimated heteroscedastic variance (inverted samples from the parameterized Gamma) and some pre-defined, larger, non-trainable variance. They perform ablation and find that their methods are complementary with the locality sampler and Student's $t$ distribution individually providing the most benefit. We use \textbf{Detlefsen} to refer to their top method, which employs all four of their proposals and generally outperforms their chosen baselines: Gaussian processes \citep{williams2006gaussian,snelson2006sparse,damianou2013deep}, unmodified neural-network parameterizations of mean and variance \citep{nix1994estimating,bishop1994mixture,kingma2013auto,rezende2014stochastic}, Bayesian neural networks \citep{mackay1992practical,hernandez2015probabilistic}, and Monte-Carlo Drop Out \citep{gal2016dropout}. We independently implemented just their Student's $t$ proposal and refer to it as \textbf{Student}. This proposal in isolation is an important baseline since our variational methods also produce a Student's $t$ posterior predictive $p(y|x)$.
 
\subsection{Proposed Regression Methods}
\label{sec:variational-variance}
In contrast to \cite{detlefsen2019reliable}, we propose a single, simple modification: treat precision variationally (\cref{fig:regression}, right). The Student's $t$ variance can be undefined and arbitrarily close to $\infty$, which makes it famously robust against outliers, but, as we will see, unfortunately hamstrings its ability generate sensible data under our PPC framework. Depending on its parameters, a Gamma prior over precision can saturate in a single- (effectively lower bounding variance) or double-sided (upper and lower bounding variance) manner. Thus, we can avoid optimization instabilities while also regularizing variance to pass our PPCs.

Employing amortized VI may appear as a superfluous inference procedure since the exact posterior, $p(\lambda|x,y)$, is available (see supplement), however, it factors into $\prod_{i=1}^N p(\lambda_i|x_i,y_i)$ such that local precision depends on both $x_i$ and $y_i$. This undesirable dual dependence could mean $p(\lambda|y_i,x)\neq p(\lambda|y_j,x)$ when $y_i\neq y_j$. Thus, the exact posterior falls outside the scope of heteroscedasticity, where any realization of $x$ should surjectively map onto variance or its distribution's parameter space. Our variational family, $q(\lambda_i|x_i)\triangleq \Gam(\lambda_i|\alpha(x_i),\beta(x_i))$ satisfies this requirement. Amortized VI preserves the modeling capacity of the Student's $t$ regression \citep{detlefsen2019reliable} as it requires the same number of neural parameterizations and too yields a Student's $t$ posterior predictive. To summarize, we give up posterior exactness for heteroscedasticity and the ability to probabilistically regularize precision's variational distribution.

We employ black-box VI \citep{ranganath2014black} in conjunction with reparameterization gradients \citep{salimans2013fixed,kingma2013auto,rezende2014stochastic,figurnov2018implicit}  to maximize our variational objective,
\begin{align}
	\ELBO =
		& \sum_{(x,y)\in\D}\E_{q(\lambda|\alpha(x),\beta(x))}\Big[\log \N(y|\mu(x),\lambda)\Big] \nonumber \\
		&- D_{KL}\big(q(\lambda|\alpha(x),\beta(x)) \ || \ p(\lambda)\big)\label{eq:elbo-reg},
\end{align}
w.r.t. the networks $\mu(\cdot)$, $\alpha(\cdot)$, and $\beta(\cdot)$. The first expectation of \cref{eq:elbo-reg} conveniently evaluates analytically
\begin{align*}
    \frac{1}{2}\Big(\psi(\alpha(x)) - \log \beta(x) - \log(2\pi) -  \frac{\alpha(x)}{\beta(x)}(y - \mu(x))^2 \Big)
\end{align*}
($\psi(\cdot)$ is the Digamma function) for univariate $y$ and, with a diagonal covariance assumption, for multivariate $y$. Networks $\alpha(\cdot)$ and $\beta(\cdot)$ employ softplus activations to ensure they give positive parameter values.

\subsubsection{Precision Priors}
\label{sec:priors}
Typically, one defines a generative process before specifying inference methods, yet here, we did the reverse since we seek a probabilistically principled way to regularize a distribution over precision that depends solely on the covariates: $q(\lambda|\alpha(x),\beta(x))$, our variational posterior. As such, we still must define the priors whose regularization effects we wish to evaluate. We consider both homoscedastic priors of form $p(\lambda)$ as well as heteroscedastic priors of form $p(\lambda|x)$, but note we use $p(\lambda)$ as in \cref{eq:elbo-reg} to refer to both homo- and hetero-scedastic priors. Since heteroscedasticity is always available to our variational posterior, $q(\lambda|\alpha(x),\beta(x))$, we really only care about which prior(s) offer optimal PPC performance, but note a heteroscedastic prior creates a generative process with explicit heteroscedasticity, whereas, with a homoscedastic prior, we only recover heteroscedasticity as a result of our inference choices. Performance aside, one may find a philosophical preference to have heteroscedasticity exist congruently in both the generative process and in inference.

We begin by introducing our considered homoscedastic priors. First, we use a standard \textbf{Gamma} prior $p(\lambda)\triangleq \Gam(\lambda;a,b)$, where $a$ and $b$ are scalar parameters specified a priori. Second, we use what we call a \textbf{Variational Posterior (VAP)} prior. This prior independently sets $p(\lambda_i) \triangleq q(\lambda_i|\alpha(x_i),\beta(x_i))$ for each data point such that the KL divergence penalty in \cref{eq:elbo-reg} vanishes. This `prior' serves as an ablation test to confirm the beneficial regularization of the KL divergence. The variational objective (\cref{eq:elbo-reg}) with a VAP prior becomes a lower bound of the log predictive likelihood of the Student's $t$ regression via Jensen's inequality. We additionally consider the Empirical Bayes \textbf{VAMP} prior \citep{tomczak2017vae}, which is the prior that maximizes the ELBO: the aggregate posterior $p^*(\lambda)=N^{-1}\sum_{i=j}^{N}q(\lambda|\alpha(x_j),\beta(x_j))$, taken over the $N$ training points. We note that this summation marginalizes out heteroscedasticity. For computational efficiency, \cite{tomczak2017vae} propose using $K < N$ randomly selected (without replacement) training points (pseudo-inputs) instead of all $N$. They denote the $j$'th pseudo-input as $u_j$. Additionally, they introduce the concept of treating pseudo-inputs $\{u_1,\hdots,u_K\}$ as trainable parameters which back propagation can modify, which we denote with \textbf{VAMP$^*$}.

For heteroscedastic priors, we first consider our novel modification to the VAMP prior, \textbf{xVAMP}
\begin{align*}
    p(\lambda|x) \triangleq \sum_{j=1}^K \pi_j(x) q\big(\lambda|\alpha(u_j),\beta(u_j)\big).
\end{align*}
Heteroscedasticity is preserved using $\pi(x)$, a neural network that maps $x$ onto the simplex, to determine the mixture proportions. This augmentation decomposes the KL divergence from \cref{eq:elbo-reg} into
\begin{align}
	\E_{q(\lambda|x)}[\log q(\lambda|x)] - \E_{q(\lambda|x)}\Bigg[\log \sum_{j=1}^K \pi_j(x) q(\lambda|u_j)\Bigg],\label{eq:kl-xvamp}
\end{align}
where we evaluate the first term analytically as the Gamma distribution's negative entropy and Monte-Carlo estimate the second using the log-sum-exp trick. We derive \cref{eq:kl-xvamp} in our supplement. We too consider trainable pseudo-inputs for our xVAMP prior, which we denote as \textbf{xVAMP$^*$}. Our second heteroscedastic prior is mixture of Gamma distributions
\begin{align*}
	p(\lambda|x) \triangleq \sum_{j=1}^K \pi_j(x) \Gam(\lambda|a_j,b_j),
\end{align*}
where again the mixture proportions, $\pi(x)$, depend on the covariates via a trainable mapping onto the simplex. We denote this prior as \textbf{VBEM}, which stands for \emph{Variational Bayes Expectation Maximization}, since optimizing the prior parameters during VI is analogous to performing M steps. The resulting KL divergence is identical to \cref{eq:kl-xvamp} except that we replace $q(\lambda|\alpha(u_j),\beta(u_j))$ with $p(\lambda|a_j,b_j)$. The non-trainable set of scalar parameters $\{a_j,b_j\}_{j=1}^K$ is the Cartesian square of a set of scalars ranging from 0.05 to 4.0 (see supplement for details). Again, we consider a version with trainable parameters, which we note as \textbf{VBEM$^*$}. Here, however, we randomly initialize trainable parameters $\{\hat{a}_1,\hat{b}_1,\hdots,\hat{a}_K,\hat{b}_K\}$ using a $\Uni([-3,3])$. To ensure valid VBEM$^*$ parameters, we apply the  softplus to these parameters (e.g. $a_j = \text{softplus}(\hat{a}_j)$). Because precision is local to each data, there is always a 1:1 ratio of likelihoods to KL divergences. Thus, the Bayesian truism that growing the data set will eventually overwhelm the prior does not apply here. 

\begin{figure}[t]
    \centering
    \includegraphics[width=0.48\textwidth]{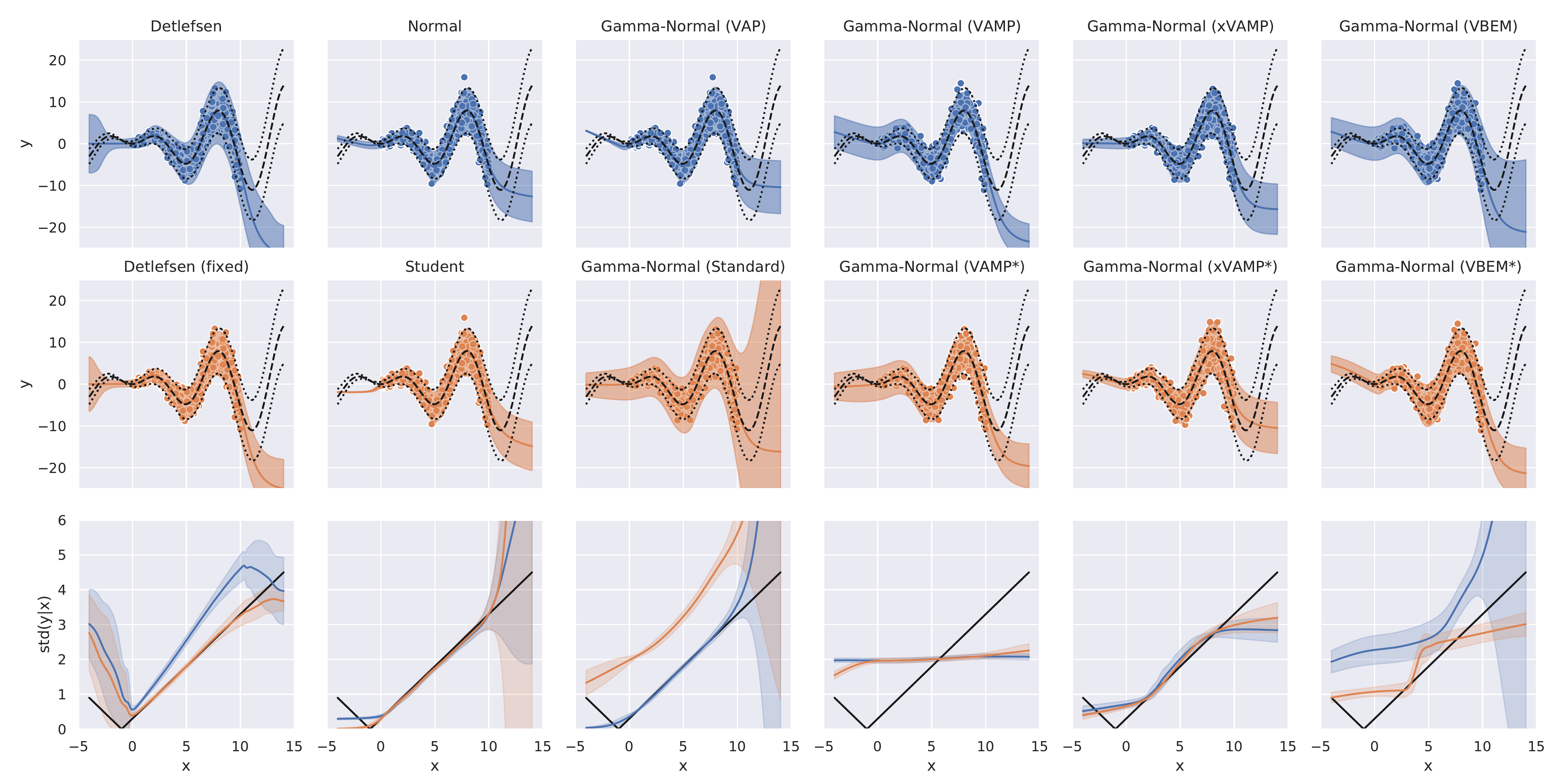}
    \caption{Toy Regression Results. Top two rows: dots are training data, black dashed/dotted lines and colored lines/areas are the true and predictive $\E[x|y] \pm 2 \cdot \sqrt{\var(y|x)}$, respectively. Third row: the true (black) and average predictive (colors correspond to methods above) $\sqrt{\var(y|x)}$ for 20 trials (area is one deviation).}
    \label{fig:toy}
\end{figure}

\subsection{Toy Data}
\cite{detlefsen2019reliable} simulate heteroscedastic data. Similarly, we define a toy process $y \triangleq x \cdot \sin(x) + \epsilon$ where $\epsilon|x \sim \N(0, [0.3 \cdot (1 + x)]^2)$. We sample training covariates uniformly from $[0,10]$ and test over equally spaced points in $[-4,14]$. \cite{detlefsen2019reliable}'s \DetCode's normal log likelihood's log variance term was missing a $\frac{1}{2}$ (this bug only affected this particular experiment). We ran their code with and without our correction to assess its impact. We also mimicked their implementation specifics (see supplement). We find all methods adeptly estimate the true mean on the training interval $[0,10]$ (top two rows of subplots, \cref{fig:toy}). Fixing \cite{detlefsen2019reliable}'s bug significantly improves their ability to learn the true variance on $[0,10]$. Our VAP prior is equally accurate on this interval. We set the parameters for the standard Gamma prior using the MLE parameters of a Gamma distribution fit over the the training interval's true precision values. The standard Gamma and VBEM priors similarly overestimate the true variance on parts of $[0,10]$, but are the only methods to consistently overestimate the variance outside the training interval $[0,10]$ (a desiderata of \cite{detlefsen2019reliable}). The VAMP and VAMP$^*$ priors are poor at capturing heteroscedastic variance, which we attribute to the aforementioned marginalization of heteroscedasticity. Namely, $D_{KL}(q(\lambda|x)||K^{-1}\sum_{j=1}^K q(\lambda|u_j))$ is minimized when the variational distributions are approximately uniform. Indeed, they predict homoscedasticity with a nearly constant standard deviation that is approximately equal to the expected value of the true standard deviation over the training interval, 1.8. The xVAMP and xVAMP$^*$ priors generally capture the true variance albeit with some inaccuracy near the closure of $[0,10]$. VBEM$^*$ behaves similarly but with added inaccuracy across the entire training interval.

\begin{table*}[ht]
    \caption{UCI Log Predictive Probability \meanpmstd \ \tuple}
    \label{tab:uci-ll}
    \begin{center}
	    {\scriptsize \input{assets/regression_uci_ll.tex}}
    \end{center}
\end{table*}

\subsection{UCI Data}
\label{sec:uci}
We consider many of the same \href{http://archive.ics.uci.edu/ml/index.php}{UCI} datasets as \cite{detlefsen2019reliable} and process them similarly: we independently whiten all features and targets to enforce zero mean and unit variance, while reporting performance metrics for the original target scalings. We collect metrics from randomly held-out validation sets that each constitute 10\% of the data across 20 trials. We match the remaining implementation specifics to \cite{detlefsen2019reliable} (see supplement for details).

We report  model likelihood, $N^{-1}\sum_{i=1}^N \log p(y_i|x_i)$, normalized by $N$, the number of validation data. Recall $p(y_i|x_i)$, the (posterior) predictive likelihood, is a Student's $t$ for all methods except the Normal model. The remaining PPC metrics require residuals for the predictive mean $\E[y_i|x_i] - y_i$, variance $\var[y_i|x_i] - \big(\E[y_i|x_i] - y_i\big)^2 $, and samples $\big(y^{*} \sim p(y_i|x_i)\big) - y_i$. We note the expectations and variances are w.r.t. $p(y_i|x_i)$. The predictive variance is $\sigma^2(x)$ for the Normal model and $\forall \alpha(x) > 1: \frac{\beta(x)}{\alpha(x) - 1}$ (i.e. the expectation of an Inverse-Gamma), which is always available since we offset $\alpha(x)$'s softplus output by 1. For each residual, we compute bias (i.e. the mean) and root mean square error (RMSE) over the $N$ validation points.

\Cref{tab:uci-ll} contains UCI log predictive probabilities with top performers in bold. We include recent competitive results \citep{sun2019functional}, which we italicize to emphasize that they are reported--we did not reproduce them. Often the reported models utilized larger neural networks than we did. If multiple architectures were reported, we selected the one closest in size to ours. We report tables corresponding to the remaining six PPC metrics in our supplement, though we include a summary in \cref{tab:uci-cc}, which tallies the number of the datasets for which a method was the top performer and, in parentheses, was statistically indistinguishable from the winner according to a two-sided Kolmogorov–Smirnov test with a $p\leq 0.05$ significance threshold. Because we could not obtain all PPC metrics for reported results, we exclude them from the summary table, but append them to the corresponding PPC table in our supplement if they were available.

Examining \cref{tab:uci-cc}, we find that our baselines generally under perform in each category compared to our methods. Student's and VAP's under performance confirms the benefit of regularizing variance. Our VBEM$^*$ prior offers the best balance of performance, winning log likelihood and posting a competitive number of statistical ties for the remaining PPCs. Since particular applications may place higher emphasis on certain PPC categories, we encourage readers to utilize \cref{tab:uci-cc} as a guide when picking a method most appropriate for their application.

\begin{table*}[t]
    \caption{UCI Regression Summary: reported as wins (statistical ties)}
    \label{tab:uci-cc}
    \begin{center}
	    {\scriptsize \input{assets/regression_uci_champions_club.tex}}
    \end{center}
\end{table*}

\section{Variational Variance for VAEs}
\label{sec:vae}
The variational autoencoder (VAE) \citep{kingma2013auto} is a deep latent variable model (DLVM) that provides computationally efficient VI for a generative process from a low-dimensional latent local variable $z_i$ to high-dimensional data $x_i$. We place a $p(z)\triangleq\N(0,I)$ prior on the latent variables and perform inference by defining $q(z|x)\triangleq \N(z|\mu_z(x),\sigma_z^2(x))$, where $\mu_z(x)$ and $\sigma_z^2(x)$ are bifurcated outputs of the same neural network. A softplus is applied to the variances to ensure positivity. The VAE's ELBO is
\begin{align}
    \sum_{x\in\D} & \E_{q(z|x)}\Big[\log \N(x|\mu_x(z),\sigma^2_x(z))\Big] \nonumber \\ &- D_{KL}\big(q(z|x) \ || \ p(z)\big)\label{eq:elbo-vae}.
\end{align}
Parameter maps $\mu_x(z)$ and $\sigma^2_x(z)$ can be either bifurcated outputs of the same neural network (\textbf{VAE}) or 
separate neural networks (\textbf{VAE-Split}). We evaluate both architectures with and without batch normalization (\textbf{+ BN}). \cite{detlefsen2019reliable} apply their regression proposals to VAEs. We refer the interested reader to their manuscript for details. \cite{takahashi2018student} propose using a Student's $t$ likelihood to bolster optimization and improve model likelihood. Their method, \textbf{VAE-Student}, results in an ELBO
\begin{align}
    \sum_{x\in\D} & \E_{q(z|x)}\Big[\log \T(x|\mu_x(z),\lambda_x(z),\nu_x(z))\Big] \nonumber \\ &- D_{KL}\big(q(z|x) \ || \ p(z)\big)\label{eq:elbo-vae-student}
\end{align}
with three separate neural networks, $\mu_x(z)$, $\lambda_x(z)$ and $\nu_x(z)$ for mean, precision, and degrees-of-freedom, respectively. Since the Student's $t$ variance is undefined for $\nu_x(z)\in(0,1]$, infinite for $\nu_x(z)\in(1,2]$, and arbitrarily close to $\infty$ for $\nu_x(z)\approx 2$, we restrict $\nu_x(z) > 3$ using a shifted softplus. We found that allowing the posterior predictive to attain these high variances worsens its PPC performance beyond what we report. \cite{takahashi2018student} additionally propose their \textbf{MAP-VAE} where precision is absorbed into the likelihood: $p(\lambda|z)\triangleq \Gam(\lambda|a,b)$ for pre-defined constants $a$ and $b$. The MAP-VAE's ELBO is identical to \cref{eq:elbo-vae} except for the additional log likelihood term and replacing $\sigma^2_x(z)$ with a network that outputs precision. Our method, \textbf{V3AE} (variational variance VAE) treats precision variationally and uses $q(\lambda|z)\triangleq \Gam(\lambda|\alpha(z),\beta(z))$ as its variational family. We consider the same priors discussed in \cref{sec:priors} for $\lambda$, except that we now condition on latent codes $z_i$. Our resulting ELBO,
\begin{align}
    \sum_{x\in\D}&\E_{q(z|x)}\Bigg[
        \E_{q(\lambda|z)}\Big[\log \N(x|\mu_x(z),\lambda)\Big] \nonumber \\ &-
        D_{KL}\big(q(\lambda|\alpha(z),\beta(z)) \ || \ p(\lambda)\big)\Bigg] \nonumber \\ &-
        D_{KL}\big(q(z|x) \ || \ p(z)\big)\label{eq:elbo-v3ae},
\end{align}
introduces a KL divergence that regularizes the predictive variance. See supplement for additional details.

A posterior predictive distribution is always the expected likelihood w.r.t. the (variational) posterior. The VAE (+ BN), VAE-Split (+ BN), and MAP-VAE's predictive distribution
is the expected normal likelihood w.r.t. $q(z|x)$. Because the decoder employs neural networks that operate on $z$, this integral is not analytically available. We therefore estimate it with 20 Monte-Carlo samples. The resulting approximation becomes a uniform mixture of Gaussians with one component for each $z$ sample. Similarly, the VAE-Student's approximate posterior predictive is a uniform mixture of Student's $t$ distributions. Our V3AE methods have two variational distributions $q(z|x)$ and $q(\lambda|z)$. Integrating the V3AE's normal likelihood w.r.t. $q(\lambda|z)$ is analytically tractable and gives back a Student's $t$. Thereafter, we approximate $q(z|x)$ integration again by taking a uniform mixture over $z$ samples of Student $t$ distributions. 

\begin{table*}[ht]
    \caption{VAE PPCs for Fashion MNIST \meanpmstd}
    \label{tab:vae-fashion}
    \begin{center}
	    {\scriptsize \input{assets/generative_table_fashion_mnist.tex}}
    \end{center}
\end{table*}

Having explicated the posterior predictive approximations for each VAE method, we can now define our PPCs. We report  normalized model likelihood, $N^{-1}\sum_{i=1}^N \log p(x_i^*|x_i)$, for which $x_i^*$ denotes a replicated $x_i$ (i.e. $x_i^*\equiv x_i$). Our other PPCs require residuals for the predictive mean: $\E[x_i^*|x_i] - x_i$, variance: $\var[x_i^*|x_i] - (\E[x_i^*|x_i] - x_i)^2 $, and samples: $\big(x^{*} \sim p(x_i^*|x_i)\big) - x_i$. We rely on TensorFlow's mixture distribution support for generating the log likelihoods, means, variances, and samples associate with our posterior predictive approximations. Here, we focus on mean/sample RMSE and variance bias.

\begin{figure}[h!]
    \centering
    \subfloat{\includegraphics[width=0.23\textwidth]{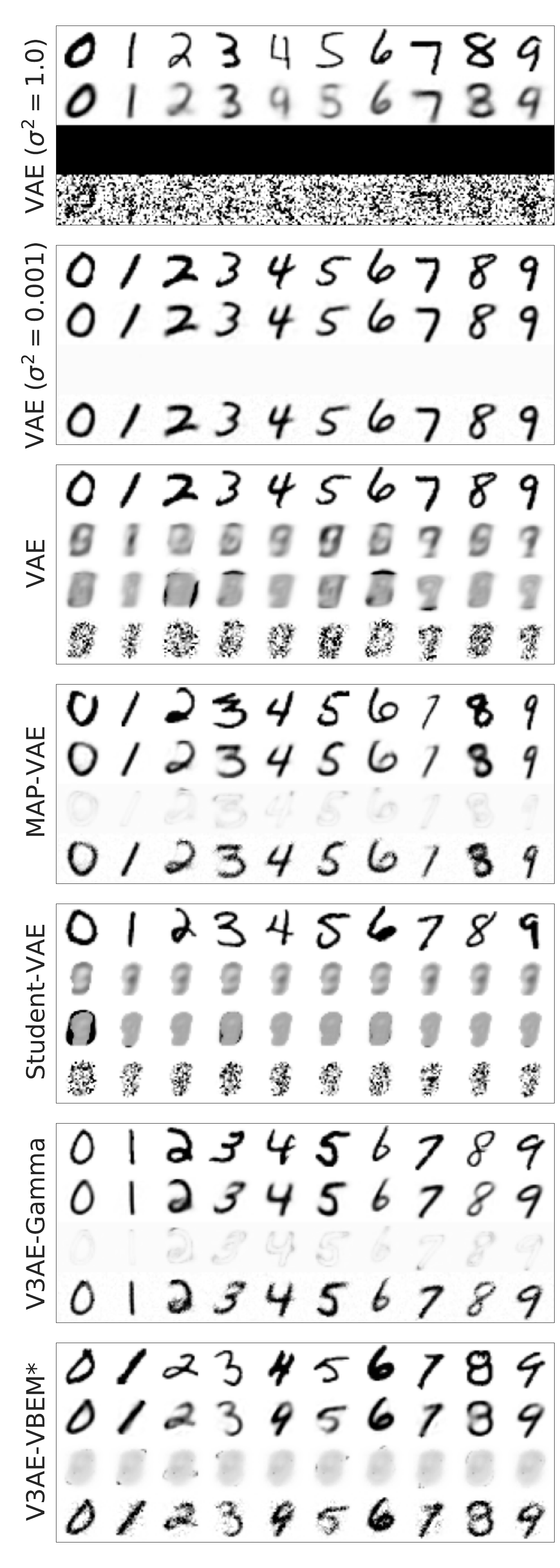}}%
    \quad
    \subfloat{\includegraphics[width=0.23\textwidth]{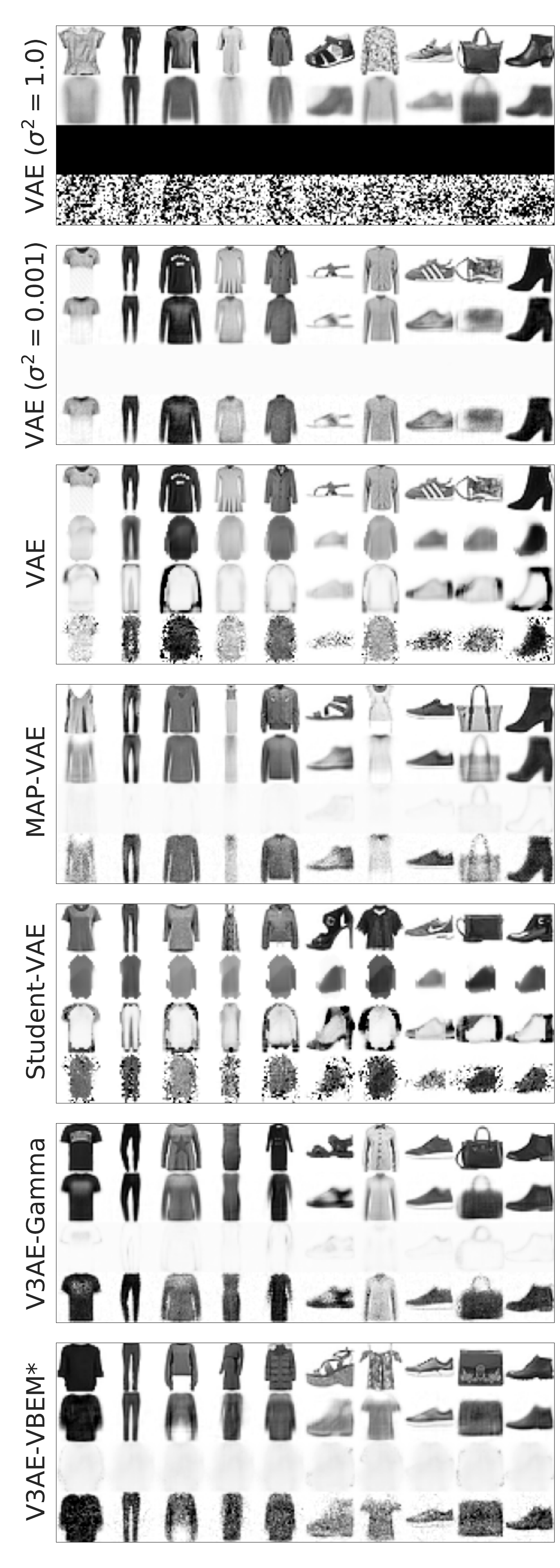}}
    \caption{VAE PPC Visualization: The rows within a subplot from top to bottom are randomly selected test data followed by the posterior predictive mean and variance and a sample from it. Pixel values are clamped to $[0,1]$, when PPC values exit this interval.}
    \label{fig:vae}
\end{figure}

We report VAE PPC metrics for the Fashion MNIST dataset in \cref{tab:vae-fashion}. Top performers are in bold as well as any method that is statistically indistinguishable using the same test from \cref{sec:uci}. In \cref{fig:vae}, we curate a subset of the VAE methods to qualitatively visualize our PPCs. We include tabular results for MNIST in our supplement, but note they trend similarly. Therein, one can also find similar figures but for all methods with additional data samples.

Comparing the Fixed-Variance VAEs confirms \cite{dai2018diagnosing}'s claim that variance impacts mean quality. While fixing variance to 0.001 produces the crispest means and samples, we sacrifice tremendous likelihood, heteroscedasticity, and our ability to analyze model uncertainty. The Student VAE dominates likelihood, but fails terribly at our other PPCs rendering predictive means and samples unrecognizable. This paradox highlights the need for our additional PPCs. The standard VAEs with heteroscedastic variances attain likelihoods similar to many of our methods, but perform notably worse than those same methods on the other PPCs. Detlefsen's VAE has the worst variance calibration and sample quality because their variance extrapolation learns to fully weight the larger constant variance of 10, which is excessive for data in $[0,1]$. The MAP-VAE and our V3AE-Gamma make up much of likelihood lost when fixing variance to 0.001. These two methods also produce crisp samples and well-calibrated predictive means and variances. We note, however, our V3AE-Gamma attains superior likelihoods. Interestingly, these two methods learn predictive variances that indicate the models are most uncertain at edge localization.

%\subsubsection*{Acknowledgements}
%All acknowledgments go at the end of the paper, including thanks to reviewers who gave useful comments, to colleagues who contributed to the ideas, and to funding agencies and corporate sponsors that provided financial support. 
%To preserve the anonymity, please include acknowledgments \emph{only} in the camera-ready papers.

\section{Conclusion}
In this manuscript, we motivate and advocate a probabilistically principled method for regularizing neural network variance parameterizations that broadly applies to regression and continuous VAEs. Our experiments highlight that model likelihood is not necessarily the complete story. Baseline methods with high likelihoods can have poor predictive mean and variance calibrations and also exhibit systematic sampling discrepancies. As we discuss, these undesirable characteristics unfortunately are indicative of model failure. Conversely, our proposed methods boast competitive model likelihoods and improve performance under this holistic set of model critiques. Additionally, our methods preserve heteroscedasticity and thereby enable model uncertainty analyses. Thus, our methods may be of interest to those working on reinforcement learning, active learning, and data imputation tasks.

\bibliography{references}

\supplement

\end{document}

%% file: assets/regression_uci_ll.tex
\begin{tabular}{lllllll}
\toprule
             &       &                   boston &                   carbon &                 concrete &                  energy &                    naval \\
Algorithm & Prior& (506, 13, 1)& (10721, 5, 3)& (1030, 8, 1)& (768, 8, 2)& (11934, 16, 2)\\
\midrule
\cite{sun2019functional} & N/A & \textit{-2.30$\pm$0.04} & -- & \textit{-3.10$\pm$0.02} & \textit{-0.68$\pm$0.02} & \textit{7.13$\pm$0.02} \\
Detlefsen & N/A &           -2.98$\pm$0.09 &            8.77$\pm$0.24 &           -3.66$\pm$0.08 &          -4.90$\pm$0.27 &            9.67$\pm$0.19 \\
Normal & N/A &           -2.42$\pm$0.23 &           13.20$\pm$1.35 &           -3.06$\pm$0.17 &          -0.48$\pm$0.69 &           14.15$\pm$0.17 \\
Student & N/A &           -2.37$\pm$0.19 &  \textbf{17.19$\pm$0.21} &           -3.10$\pm$0.17 &           0.22$\pm$0.31 &           13.60$\pm$0.39 \\
Gamma-Normal & VAP &           -2.36$\pm$0.17 &           15.52$\pm$0.24 &           -3.12$\pm$0.17 &           0.17$\pm$0.44 &           13.36$\pm$0.41 \\
             & Standard &           -2.48$\pm$0.29 &           11.28$\pm$0.02 &           -3.20$\pm$0.16 &          -1.05$\pm$0.18 &           12.33$\pm$0.16 \\
             & VAMP &           -2.39$\pm$0.17 &           14.37$\pm$0.17 &           -3.09$\pm$0.16 &          -0.18$\pm$0.21 &           14.16$\pm$0.78 \\
             & VAMP* &           -2.39$\pm$0.16 &           14.38$\pm$0.12 &           -3.09$\pm$0.16 &          -0.16$\pm$0.20 &           13.96$\pm$0.88 \\
             & xVAMP &  \textbf{-2.33$\pm$0.17} &           15.38$\pm$0.24 &           -3.01$\pm$0.14 &           0.05$\pm$0.28 &           13.50$\pm$0.59 \\
             & xVAMP* &           -2.33$\pm$0.17 &           15.41$\pm$0.18 &           -3.01$\pm$0.13 &           0.11$\pm$0.39 &           13.34$\pm$0.47 \\
             & VBEM &           -2.46$\pm$0.11 &            4.57$\pm$1.00 &           -3.11$\pm$0.07 &          -4.52$\pm$0.26 &            9.02$\pm$0.61 \\
             & VBEM* &           -2.36$\pm$0.14 &           14.64$\pm$0.16 &  \textbf{-2.99$\pm$0.13} &  \textbf{0.49$\pm$0.28} &  \textbf{14.42$\pm$0.15} \\
\midrule
             &       &              power plant &        superconductivity &                 wine-red &               wine-white &                    yacht \\
Algorithm & Prior& (9568, 4, 1)& (21263, 81, 1)& (1599, 11, 1)& (4898, 11, 1)& (308, 6, 1)\\
\midrule
\cite{sun2019functional} & N/A & \textit{-2.83$\pm$0.01} & -- & -- & -- & \textit{-1.03$\pm$0.03} \\
Detlefsen & N/A &        -3.26$\pm$9.1e-03 &           -5.21$\pm$0.02 &           -1.04$\pm$0.06 &           -1.12$\pm$0.04 &           -3.15$\pm$0.10 \\
Normal & N/A &           -2.82$\pm$0.05 &           -3.51$\pm$0.10 &           -0.92$\pm$0.05 &           -1.05$\pm$0.04 &           -1.55$\pm$0.65 \\
Student & N/A &  \textbf{-2.78$\pm$0.03} &           -3.41$\pm$0.05 &  \textbf{-0.80$\pm$0.10} &           -1.05$\pm$0.04 &           -1.73$\pm$0.59 \\
Gamma-Normal & VAP &           -2.81$\pm$0.04 &           -3.45$\pm$0.06 &           -0.87$\pm$0.06 &           -1.04$\pm$0.04 &           -1.79$\pm$0.50 \\
             & Standard &           -2.88$\pm$0.03 &           -3.45$\pm$0.04 &           -0.98$\pm$0.07 &           -1.13$\pm$0.05 &           -1.73$\pm$0.38 \\
             & VAMP &           -2.83$\pm$0.03 &           -3.94$\pm$0.02 &           -0.94$\pm$0.05 &           -1.05$\pm$0.04 &           -2.83$\pm$0.70 \\
             & VAMP* &           -2.83$\pm$0.03 &           -3.94$\pm$0.03 &           -0.94$\pm$0.05 &           -1.05$\pm$0.04 &           -2.77$\pm$0.77 \\
             & xVAMP &           -2.81$\pm$0.04 &           -3.40$\pm$0.04 &           -0.90$\pm$0.05 &           -1.03$\pm$0.04 &           -1.68$\pm$0.38 \\
             & xVAMP* &           -2.81$\pm$0.04 &  \textbf{-3.39$\pm$0.05} &           -0.89$\pm$0.06 &           -1.03$\pm$0.04 &           -1.71$\pm$0.47 \\
             & VBEM &           -2.89$\pm$0.05 &           -3.77$\pm$0.09 &           -0.91$\pm$0.05 &           -1.03$\pm$0.03 &           -2.64$\pm$0.23 \\
             & VBEM* &           -2.81$\pm$0.03 &           -3.41$\pm$0.04 &           -0.89$\pm$0.06 &  \textbf{-1.03$\pm$0.04} &  \textbf{-1.11$\pm$0.57} \\
\bottomrule
\end{tabular}

%% file: assets/regression_uci_champions_club.tex
\begin{tabular}{lllllllll}
\toprule
             &       &                       LL &                 Mean Bias &                 Mean RMSE &        Var Bias &                 Var RMSE &      Sample Bias &     Sample RMSE \\
Algorithm & Prior &                          &                           &                           &                 &                          &                  &                 \\
\midrule
Detlefsen & N/A &                    0 (0) &                     2 (3) &                     1 (1) &           0 (0) &                    0 (1) &            0 (4) &           0 (0) \\
Normal & N/A &                    0 (3) &                     1 (7) &                     2 (7) &           1 (6) &                    2 (7) &            2 (9) &  \textbf{3} (6) \\
Student & N/A &           3 (\textbf{7}) &                     0 (7) &                     0 (6) &           0 (5) &                    0 (3) &            0 (9) &           2 (5) \\
Gamma-Normal & VAP &                    0 (4) &                     0 (9) &                     0 (6) &           0 (7) &                    0 (4) &   \textbf{3} (9) &           0 (5) \\
             & Standard &                    0 (0) &                     0 (7) &                     0 (6) &  \textbf{2} (6) &                    0 (5) &            0 (7) &           0 (5) \\
             & VAMP &                    0 (3) &  \textbf{3} (\textbf{10}) &                     0 (8) &           1 (7) &  \textbf{3} (\textbf{9}) &  1 (\textbf{10}) &  1 (\textbf{9}) \\
             & VAMP* &                    0 (3) &           0 (\textbf{10}) &                     1 (8) &  \textbf{2} (7) &           1 (\textbf{9}) &  0 (\textbf{10}) &  2 (\textbf{9}) \\
             & xVAMP &                    1 (4) &                     2 (9) &                     0 (7) &           1 (6) &                    1 (7) &            0 (9) &           2 (6) \\
             & xVAMP* &                    1 (4) &                     1 (9) &                     0 (7) &  1 (\textbf{8}) &                    0 (8) &            1 (9) &           0 (6) \\
             & VBEM &                    0 (1) &           0 (\textbf{10}) &  \textbf{5} (\textbf{10}) &           0 (0) &                    2 (6) &            1 (9) &           0 (0) \\
             & VBEM* &  \textbf{5} (\textbf{7}) &                     1 (9) &                     1 (7) &  \textbf{2} (5) &                    1 (7) &            2 (9) &           0 (5) \\
\bottomrule
\end{tabular}

%% file: assets/generative_table_fashion_mnist.tex
\begin{tabular}{lllll}
\toprule
{} &                          LL &                     Mean RMSE &                      Var Bias &                   Sample RMSE \\
Method                 &                             &                               &                               &                               \\
\midrule
Fixed-Var. VAE (1.0)   &            -730.05$\pm$0.11 &              0.15$\pm$1.8e-03 &              0.98$\pm$5.7e-04 &              1.01$\pm$3.7e-04 \\
Fixed-Var. VAE (0.001) &           -1452.44$\pm$3.65 &  \textbf{9.4e-02$\pm$4.6e-05} &          -7.8e-03$\pm$8.6e-06 &  \textbf{9.9e-02$\pm$4.7e-05} \\
VAE                    &           2154.31$\pm$42.11 &              0.25$\pm$1.4e-03 &           3.3e-02$\pm$1.5e-03 &              0.39$\pm$3.1e-03 \\
VAE + BN               &           1639.39$\pm$15.33 &              0.20$\pm$1.8e-03 &           2.1e-02$\pm$3.0e-03 &              0.31$\pm$5.6e-03 \\
VAE-Split              &           2099.28$\pm$39.97 &              0.27$\pm$2.9e-03 &           4.7e-02$\pm$1.6e-03 &              0.45$\pm$4.8e-03 \\
VAE-Split + BN         &           1948.30$\pm$25.87 &              0.26$\pm$6.2e-03 &           3.1e-02$\pm$3.6e-03 &              0.41$\pm$1.1e-02 \\
Detlefsen              &        -1624.12$\pm$8.3e-03 &              0.16$\pm$8.2e-04 &              9.97$\pm$3.3e-04 &              3.17$\pm$1.2e-03 \\
MAP-VAE                &           1003.51$\pm$32.75 &              0.11$\pm$4.1e-03 &          -9.1e-03$\pm$6.2e-04 &              0.13$\pm$4.8e-03 \\
Student-VAE            &  \textbf{3134.52$\pm$18.60} &              0.29$\pm$3.3e-03 &           7.4e-02$\pm$1.6e-02 &              0.49$\pm$2.2e-02 \\
V3AE-VAP               &           2146.46$\pm$67.83 &              0.28$\pm$3.5e-03 &  \textbf{9.9e-04$\pm$3.0e-03} &              0.40$\pm$8.2e-03 \\
V3AE-Gamma             &           1201.95$\pm$25.25 &              0.11$\pm$2.8e-03 &          -8.0e-03$\pm$4.1e-04 &              0.12$\pm$3.4e-03 \\
V3AE-VAMP              &           1632.22$\pm$12.89 &              0.17$\pm$1.3e-03 &           1.5e-03$\pm$2.7e-04 &              0.25$\pm$2.4e-03 \\
V3AE-VAMP*             &           1630.10$\pm$17.87 &              0.18$\pm$2.6e-03 &           1.3e-03$\pm$2.5e-04 &              0.25$\pm$3.4e-03 \\
V3AE-xVAMP             &           1601.60$\pm$21.49 &              0.18$\pm$2.5e-03 &           1.3e-03$\pm$4.1e-04 &              0.25$\pm$3.8e-03 \\
V3AE-xVAMP*            &           1619.97$\pm$25.95 &              0.18$\pm$3.3e-03 &           1.5e-03$\pm$4.6e-04 &              0.25$\pm$5.0e-03 \\
V3AE-VBEM              &             306.46$\pm$1.04 &              0.10$\pm$6.8e-04 &           6.4e-02$\pm$9.7e-05 &              0.29$\pm$3.3e-04 \\
V3AE-VBEM*             &            1153.11$\pm$4.20 &              0.10$\pm$5.5e-04 &  \textbf{4.4e-04$\pm$3.9e-05} &              0.15$\pm$8.0e-04 \\
\bottomrule
\end{tabular}